\documentclass[letterpaper, 10 pt, conference]{ieeeconf}

\IEEEoverridecommandlockouts
\usepackage{kantlipsum}
\usepackage[english]{babel}

\usepackage{amsmath}
\usepackage{bbm}
\usepackage{dsfont}
\usepackage[colorlinks=true, allcolors=blue]{hyperref}
\usepackage{comment}
\usepackage[dvipsnames]{xcolor}

\usepackage[font=small]{caption}

\usepackage{algorithm}
\usepackage{algpseudocode}

\usepackage{spverbatim}

\newtheorem{Problem}{Problem}

\usepackage{graphicx}
\usepackage{etoolbox}
\newcommand{\insertfig}{\includegraphics[width=\linewidth]{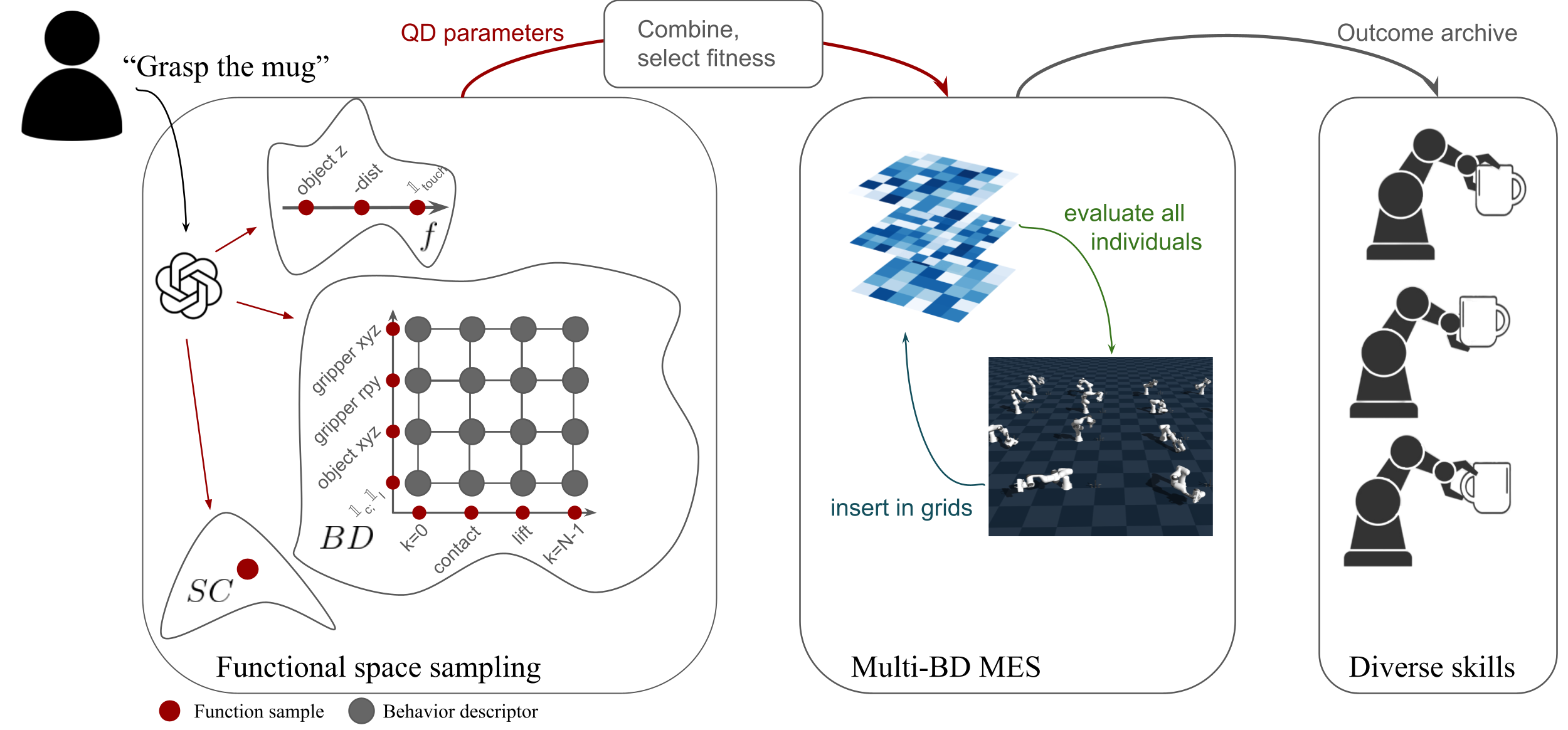}\captionof{figure}{\textbf{Overview of the proposed approach:} (left) After the user inputs the task description, samples from the behavior descriptor (BD), fitness (f) and success condition (SC) spaces are inferred with an LLM (see the appendix for a full description of the function samples in the figure). The samples are all combined, yielding three multi-BD archives. (centre) The archive with the highest-performing fitness function is kept and trained using a multi-BD MAP-Elites Success algorithm. (right) This yields an archive of diverse robotic motions, able to adapt on-the-fly to unseen task instances.}}

\makeatletter
\apptocmd{\@maketitle}{\centering\insertfig}{}{}
\makeatother

\begin{document}

\title{Autonomously Acquiring Robot Manipulation Skills with Language-Driven Quality-Diversity}
\author{\textbf{Émiland Garrabé, Mahdi Khoramshahi, Stéphane Doncieux}$^\ast$
\thanks{$^\ast$ \textbf{All authors with ISIR, Sorbonne Université, France}. Code available at \url{https://tinyurl.com/mvyjs3d5}}
}
\maketitle
\begin{abstract}
Quality-diversity (QD) algorithms have been gaining traction in robot learning, where diverse motion primitive libraries allow robots to adapt zero-shot to constraints at deployment time. However, such methods typically require expert designers to write the success condition, fitness and diversity metrics, and this strongly limits the robot's autonomy. On the other hand, existing LLM-based reward-shaping techniques allow robots to learn autonomously but only output single high-performing solutions, limiting the robot's adaptability. In this paper, we propose an approach designed to output diverse motion primitive archives by autonomously leveraging quality-diversity algorithms, only requiring a free-form description of the task in common language. To address the difficulty of designing relevant fitness and diversity metrics, we propose an autonomous exploration mechanism able to reliably output sets of functionals covering the fitness and behavior descriptor (BD) space. First, we pose policy exploration as a functional design problem, where the functional spaces are lower-dimensional than the full BD and fitness spaces, and propose an LLM-based exploration scheme to sample from these low-dimensional spaces without any task-specific prompts, fine-tuning or expert intervention. We adapt a multi-BD variant of the MAP-Elites success (MES) algorithm, designed to leverage the heterogeneous BD samples. Finally, through experiments based on the genesis simulator, we show that our method effectively generates archives of diverse motion primitives, outperforming classical QD algorithms with inferred and hand-written parametrizations on a set of $4$ robotic manipulation tasks.
\end{abstract}

\section{Introduction}
LLMs are rapidly emerging as a key tool for robotics, where their compatibility with free-form instructions and closed-syntax coding languages allows them to bridge user-robot semantic gaps. The world knowledge embedded in their large training datasets also affords them a moderate degree of independence, leading to encouraging results in open-ended robotics. While early works focused on high-level problems such as semantic task planning \cite{llmpp, cap}, a growing trend envisions the possibility of acquiring robotic skills by using LLMs as \textit{reward designers} \cite{l2r, eureka}. Leveraging the LLM's ability to write code based on open instructions, such works propose pipelines to infer reward code to learn novel skills.\\
Another powerful tool in robotics is the family of so-called quality-diversity (QD) evolutionary algorithms \cite{ME, ME_nature}. Such algorithms seek to ensure the \textit{diversity} of candidate solutions w.r.t. one another, and this leads to more efficient exploration of non-convex fitness landscapes. In robotics, this also allows designers to obtain archives of motion primitives that achieve a task in diverse ways, providing useful robustness to unseen task requirements \cite{affordance}.\\
From the point of view of autonomous robotics, one drawback of QD techniques is that they rely on carefully curated fitness signals, diversity metrics and success conditions. In practice, this means that for each new skill, an expert designer needs to write these functions by hand, usually through trial-and-error. While some works \cite{vqe} propose to autonomously learn diversity metrics through unsupervised learning, this reduces explainability, as the learned representations emerge from the learning process without explicitly taking the task into account.\\
In this paper, we explore the use of LLMs to design QD algorithms able to learn diverse robot trajectories given a task. Specifically, our contributions are: (i) we \textbf{formalize the problem of parameterizing QD algorithms to generate an archive of robotic motion primitives for a task} as a functional exploration problem; (ii) we propose an \textbf{LLM-based sampling technique}, designed to explore the success condition, fitness and BD spaces without requiring task-specific prompts, fine-tuning, few-shot examples or expert input; (iii) we propose a technique to \textbf{leverage the inferred samples} (later referred to as \textit{BD mesh}) \textbf{to generate diverse trajectory data} using a multi-BD variant of the MES algorithm and (iv) through numerical examples, we show this exploration technique \textbf{outperforms classical QD algorithms} with expert-written and LLM-inferred metrics and a genetic algorithm baseline. Finally, we provide design guidelines and identify research directions for the autonomous acquisition of skills using QD techniques.

\section{Related works}
\subsection{Quality-Diversity algorithms}
QD algorithms such as MAP-Elites have met wide successes in robotics. Early research focused on locomotion tasks \cite{cully2015robots}, where robots learn diverse gaits to overcome unseen constraints. Subsequent advances have focused on evolving closed-loop policies \cite{policy_me, policy_sigaud}, while others have shown that skills obtained using QD methods can be distilled into a robust, language-conditioned controller \cite{distill1, distill2}.\\
Recently, QD algorithms have been used to approach robot manipulation tasks \cite{morel2022automatic}, where they are a useful tool for tackling the difficult exploration problem of learning grasping motions \cite{dexg}. Early works focused on using parametrized curves in cartesian space as candidate solutions \cite{qd1}, but including task-specific priors in QD search processes has led to the generation of impressively large datasets of motion primitives \cite{qd2, qd3}. Finally, in the context of grasping \cite{affordance} and navigation \cite{ME_nature}, it has been shown that motion primitive archives generated using QD algorithms lead to robustness to unseen task requirements. 

\subsection{Foundation models as reward designers}
While LLMs can be used to directly specify reward values in some settings \cite{kwon2023reward}, their code-writing ability can also be leveraged to infer reward functions, allowing designers to bridge the gap between open-semantics goal formulations and learning algorithms \cite{l2r}. This approach, coupled with curriculum approaches \cite{eureka} or planning techniques \cite{targ, supddown} has allowed robots to acquire complex skills. Many language-based reward shaping techniques propose to iteratively refine the reward code, either through direct feedback \cite{card} or through trajectory ranking \cite{zeng2024learningrewardrobotskills}.\\
While code is a popular embedding for reward functions, another noteworthy trend is that of using images as a surrogate plan for learning robot manipulation tasks through goal-conditioned RL \cite{zeroshot, surfer}, and this process can be augmented by chain-of-thought techniques \cite{cotimg}. For a thorough review of generative AI techniques for robotics, see \cite{genai}.

\subsection{Trajectory datasets in robotics}
Recent advances in machine learning allowed roboticists to design so-called Vision-Language-Action (VLA) models \cite{openvla,smolvla}, which achieved impressive performances in manipulation. A key necessity for training VLAs is the requirement for very large trajectory datasets. Current VLAs are trained by human operators, leading to high deployment costs, even if some authors mitigate this by including simulated data in the training dataset \cite{groot}.\\
While some efforts have been made to build pipelines that autonomously generate robotic data \cite{anytask}, these typically focus on RL and/or motion planning, suffering from the drawbacks discussed above.\\

Overall, most methods for autonomously acquiring robot skills focus on obtaining single, high-performing solutions. This can lead to brittleness when deployed to the real world, due to the high dimensionality of manipulation task spaces, and curricula based on these techniques need to be carefully curated. Conversely, while motion primitives obtained with QD pipelines are well-suited to out-of-the-box integration due to their diversity, the design of useful, task-specific diversity metrics and fitness signals still requires expert efforts, and this is incompatible with autonomy requirements. In this paper, we show that \textbf{LLM-based reward shaping techniques can be adapted for QD algorithms in the context of manipulation tasks}, leading to archives of diverse solutions while being compatible with free-form instructions.\\

\section{Method}\label{sec:methods}
In this section, we begin by formalizing the robotic skill archive design problem and the parametrization of a MAP-Elites-Success algorithm for robotic skill acquisition. Then, we propose a sequential technique for exploring the BD and fitness spaces using LLMs, without expert input or task-based adaptation. We finally show how the generated mesh of BD samples can be used with a multi-BD MAP-Elites-Success (MES) algorithm to generate an archive of diverse motion primitives for a given task.

\subsection{Trajectory dataset generation for robotics}
\subsubsection{Design problem}
Let the environment, robot and object be represented as a POMDP\footnote{We use the POMDP framework for generality here, but most of the POMDP assumptions are not strictly necessary.}. The robot designer specifies a \textbf{task} $t$ in language. Let $\mathcal{S}$ be the subset of the state space such that the task is achieved when the state $s$ is in $\mathcal{S}$.\\
A \textbf{task instance} $t_i$ is a subset of $S$ that induces extra requirements to be achieved by the robot. Such requirements can stem from unspecified designer needs ('Grasp the mug by the handle', 'Turn on the water at low flow') or from obstacles in the scene at deployment time, for instance.\\
We cast the motion primitive archive design problem as:
\begin{Problem}
Given a known task and unknown task instance distribution $\mathcal{T}$, we cast the robotic QD design problem as: \\
Find $A^*$ such that:
\begin{equation}\label{eq:qd_prob}
    A^* \in \underset{A \in \mathcal{A}}{\text{arg max}}\mathds{E}_{t_i \in \mathcal{T}}\left[ \mathds{1}(A \text{ achieves } t_i) \right]
\end{equation}
Where $A$ is an archive of motion primitives and $\mathcal{A}$ is the space thereof.
\end{Problem}
Intuitively, Equation \ref{eq:qd_prob} can be explained in the following way: the goal is to design the archive that is the likeliest to achieve unknown instances of a known task.\\
Traditional reward design approaches are focused on obtaining single primitives, i.e. archives that are a singleton in trajectory space. In this work, we propose to rely instead on archives of motion primitives learned using quality-diversity algorithms. It is assumed that these motion primitives can be exploited by a dedicated selection and planning process \cite{affordance} to solve the task instance and that the problem is thus to find the set of primitives that can be used to solve any instance of the given task.

\subsection{QD algorithms, MAP-Elites Success}
QD algorithms are a subset of evolutionary algorithms, where a novelty metric is used to measure how \textit{diverse} individuals are w.r.t. each other. Typically, this is done by projecting individuals in a \textit{behavior space} and measuring either distances between individuals \cite{NS} or discretizing the behavioral space and considering individuals filling separate niches in this archive as diverse \cite{cully2015robots}.\\
In this paper, we use a variant of the so-called MAP-Elites Success (MES) algorithm. In MES, which is a derivation of the archive-based MAP-Elites, a success condition is used to assess if each individual achieves the task, and successful individuals are selected in priority for evolution.\\
The main difficulty when designing MES algorithms is to find the triplet of success condition (map $\Theta \rightarrow \{0,1\}$, where $\Theta$ is the space of candidate solutions), fitness ($\Theta  \rightarrow \mathds{R}$) and behavior descriptor ($\Theta \rightarrow \mathds{R}^\infty$ with $\mathds{R}^\infty := \underset{i\geq 1}{\sqcup}\mathds{R}^i$ and $\sqcup$ is the disjointed union) that will lead to a large archive of diverse solutions. As this design problem requires expert input, we instead rely on a sampling strategy to cover the fitness and BD spaces and use the generated samples.

\subsection{LLM inference for multi-BD MES}
In this section, we introduce our strategy for parameterizing QD algorithms with LLMs.
\subsubsection{Success condition} While obtaining a correct success criterion is critically important for the resulting archive's quality, it is relatively easy to design. Accordingly, we only infer one success condition for a given task, reducing computational load.
\subsubsection{Fitness signal} In MAP-Elites Success, fitness plays a crucial role in the early stages by enabling the discovery of successful individuals. However, once a sufficient number of successful individuals have been found, fitness becomes less influential, as differences in fitness among these elites become relatively marginal. To avoid searching over the intractably large fitness space, we first assume that the fitness function can be expressed as the mean of a value computed at each simulation time step, allowing us to instead sample from the space of mappings $\mathcal{I}\rightarrow \mathds{R}$. Then, we generate $3$ fitness functions in a single LLM inference pass. The goal of this sampling without replacement is to ensure a wide coverage of the fitness space. Then, we create $3$ multi-BD archives (see lower), with each archive being assigned one of the fitness signals. After a short training run, the archive with the most successes is kept, and the two others are discarded.\\
\subsubsection{Behavior descriptor}
Despite their coding performance and 'common sense', LLMs lack expert knowledge, and this makes inferring the optimal BD zero-shot tricky. While a behavior descriptor that 'seems' relevant is easy to obtain, inferring a BD that both drives exploration in the solution space while ensuring useful diversity for the downstream task is critical.
To overcome the difficulties of this design problem, we propose to restrict the search to a subset of the BD space, and to explore it sequentially.\\
While exploring all maps $\mathcal{I}^N \rightarrow \mathds{R}^\infty$ is intractable, we propose to instead sample from the functional space $\{g\circ h: g :\mathcal{I}\rightarrow  \mathds{R}^\infty; h: \mathcal{I}^N\rightarrow \mathcal{I}\}$. Intuitively, this means we separate the choice of the information that should be used to compute the BD ($g$), and the timestep at which this information should be measured ($h$). Both of these functional spaces, being simpler, are easier to widely explore through sampling. 
While we explore the two spaces independently (i.e. with two separate inferences) to manage the context size, the candidates for $g$ should all be inferred at once (sampled without replacement) to avoid duplicates. The same goes for $h$. This BD space splitting and sampling approach is illustrated in Figure \ref{fig:method}.\\
To leverage the obtained BD mesh, we use a multi-BD variant of the MES algorithm. In this variant, a separate grid is maintained for each BD. At the evaluation step, each individual is inserted in each BD grid through local competition, as in standard MES. At the selection stage, we carry out selection normally for each grid and concatenate the resulting individual lists for mutation and evaluation. We call the concatenation of the grids linked to each BD \textit{multi-BD archives}. See Figure \ref{fig:new_qd} for a representation of the multi-BD algorithm.\\
The overall procedure is summarized in pseudocode in Algorithm \ref{alg:algo}. We also provide the prompt template for the LLM-based BD sampling mechanism in Appendix \ref{sec:ap1}.\\
\setcounter{figure}{1}

\begin{figure}
    \centering
    \includegraphics[width=0.33\linewidth]{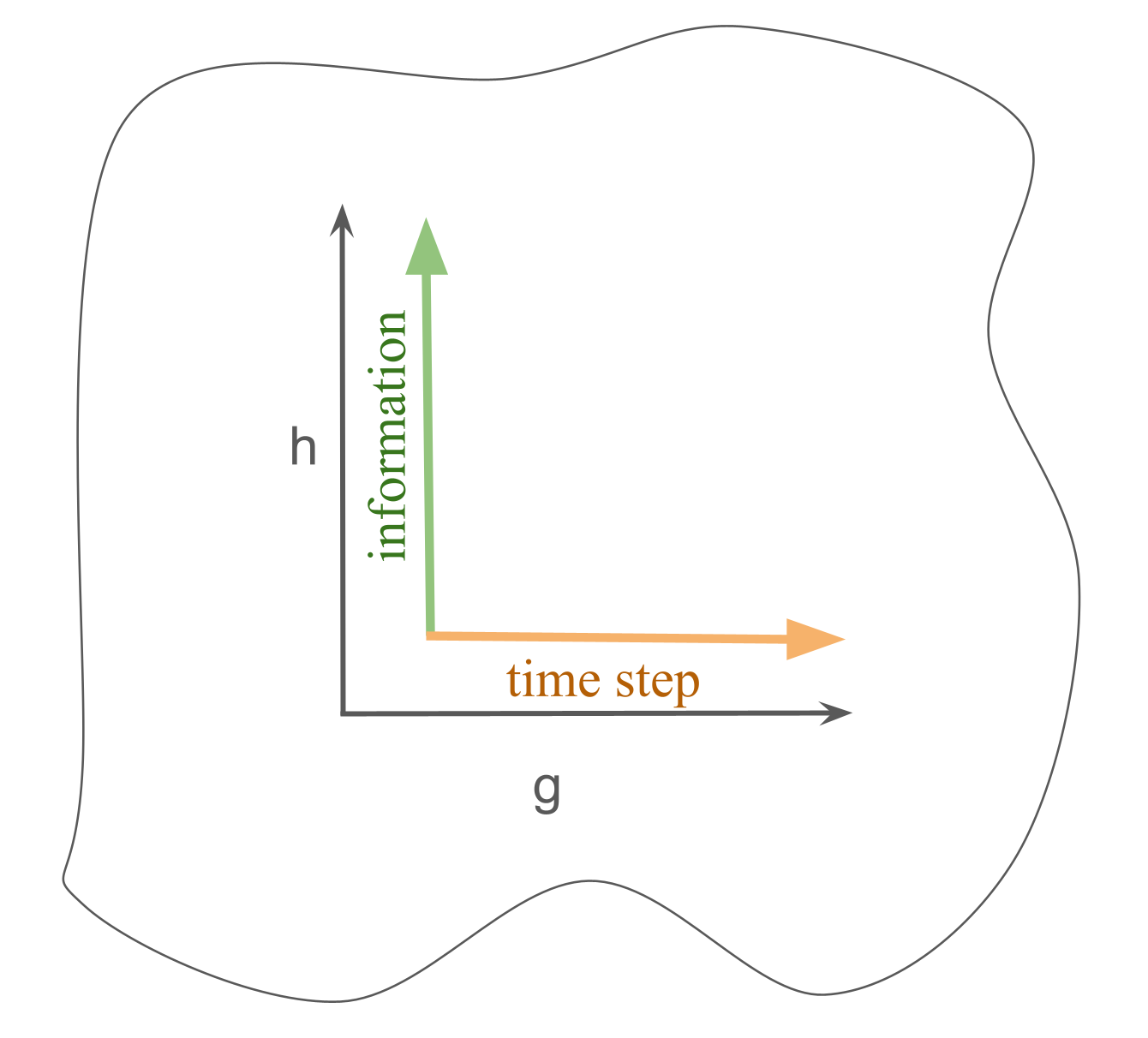}\includegraphics[width=0.33\linewidth]{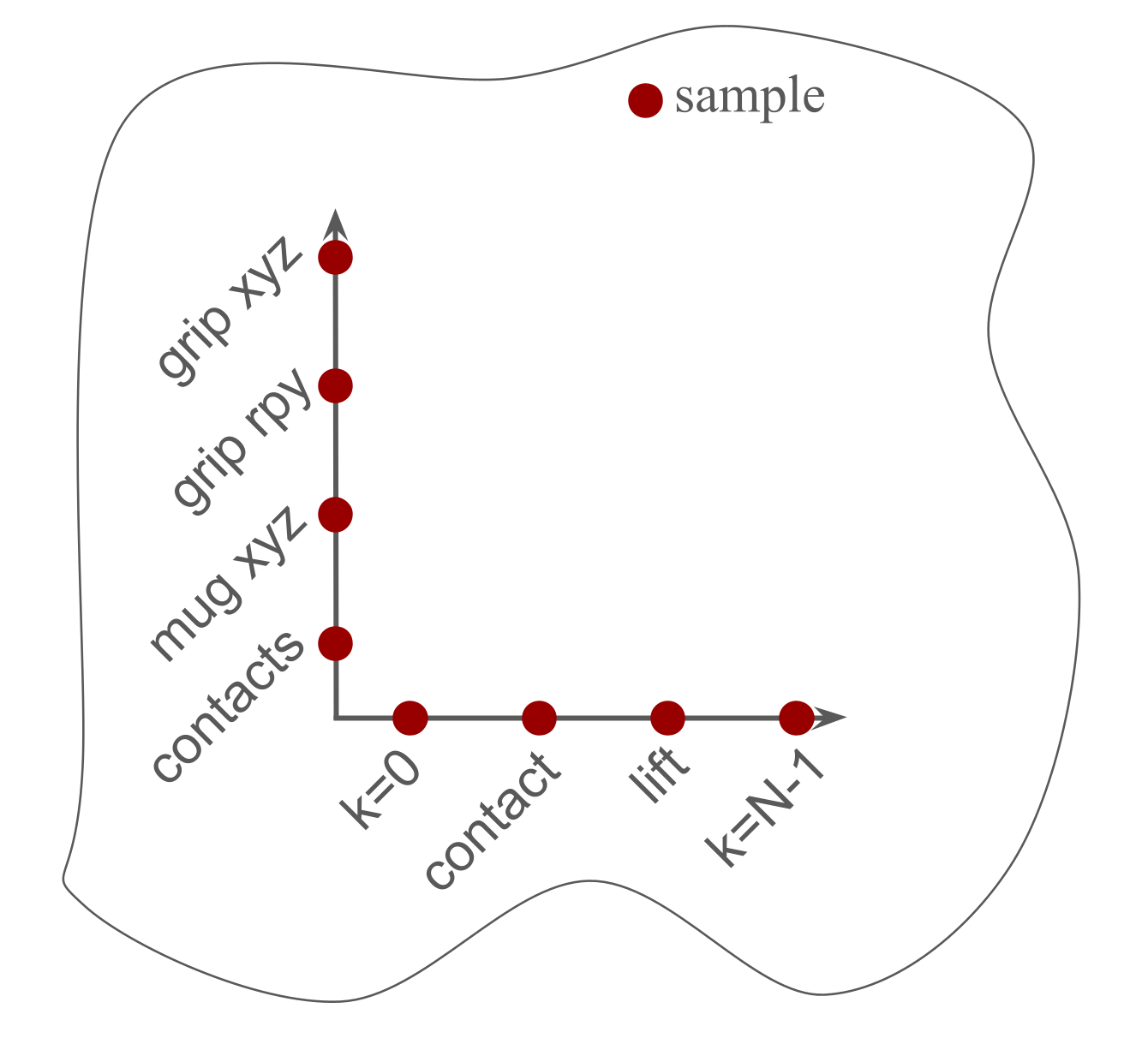}\includegraphics[width=0.33\linewidth]{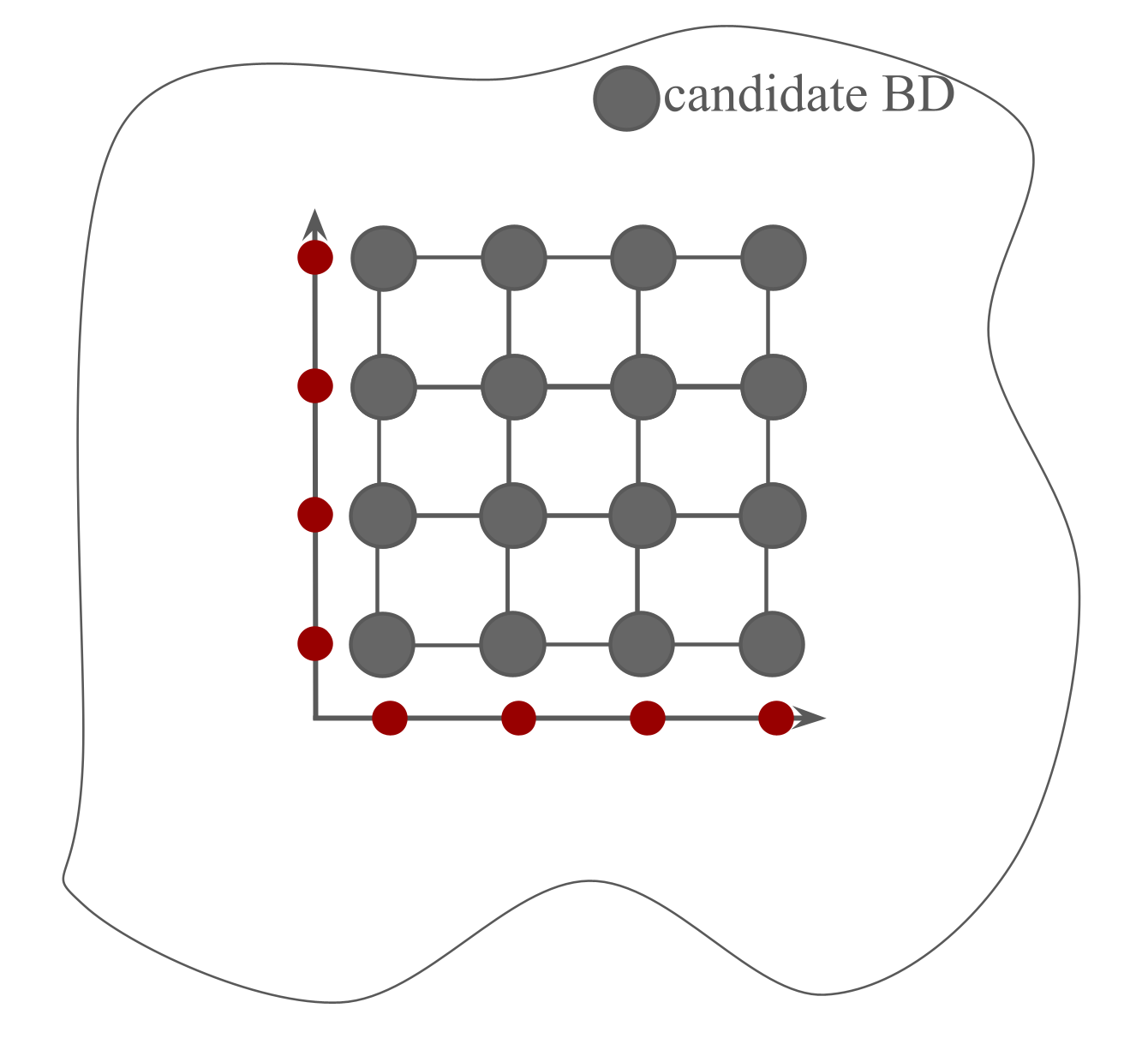}
    \caption{BD space sampling. From left to right: (i) we cast BD exploration as a two-stage process: finding the time-step and finding the information to extract; (ii) candidate functions are sampled without replacement, ensuring coverage, using an LLM (see Appendix \ref{sec:ap2} for the BD mesh corresponding to the labels); (iii) the resulting samples are composed and used as candidate BDs.}
    \label{fig:method}
\end{figure}

\begin{figure}
    \centering
    \includegraphics[width=\linewidth]{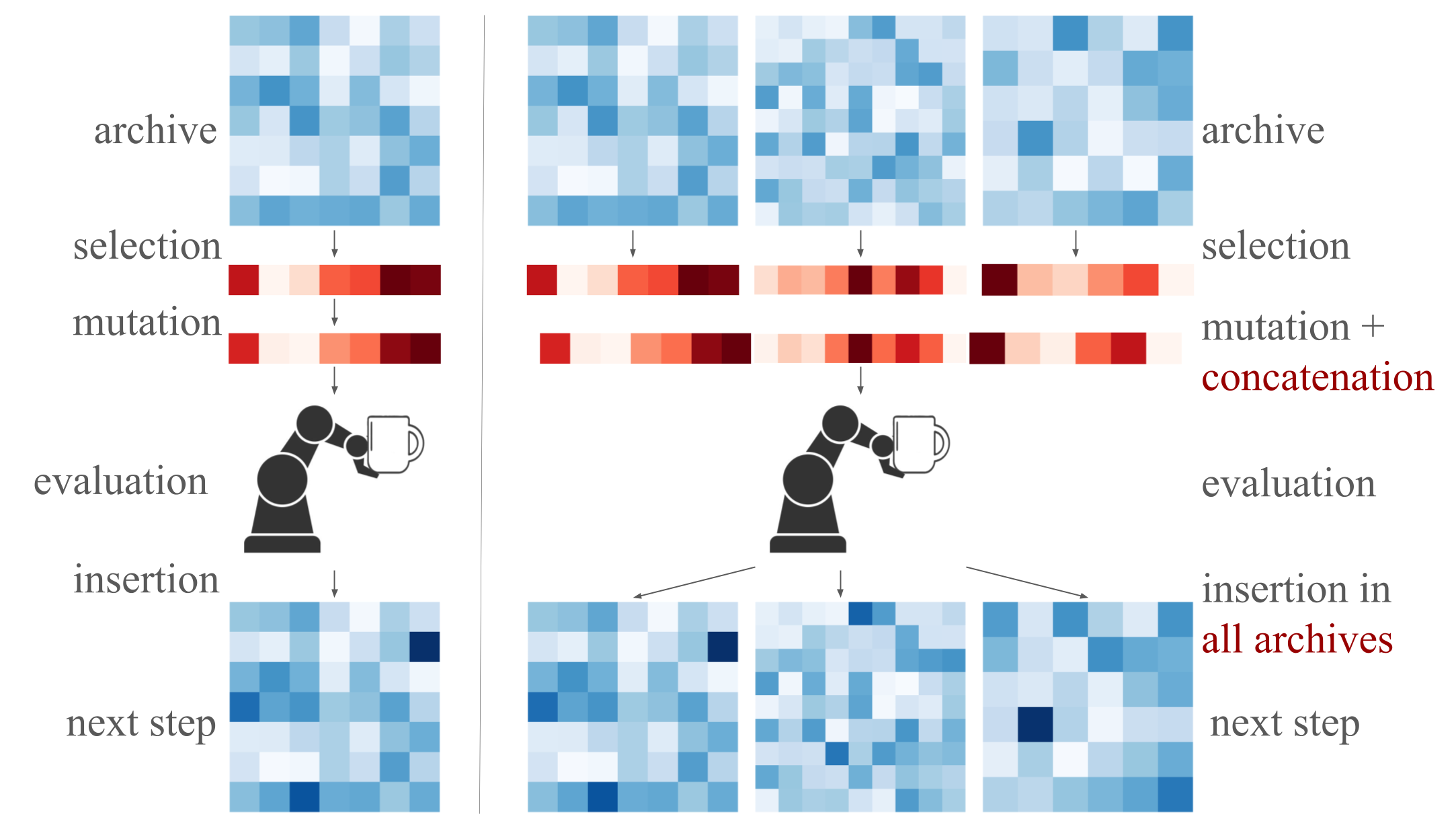}
    \caption{(left) Standard MES algorithm: individuals are selected from the archive, mutated, evaluated and inserted in the archive through local competition. (right): Multi-BD MES algorithm: after selection, individuals taken from each grid are concatenated, mutated and evaluated. The result of this evaluation then undergoes local competition in each grid of the multi-BD archive.}
    \label{fig:new_qd}
\end{figure}

\begin{algorithm}
\caption{Autonomous QD skill acquisition}\label{alg:algo}
\hspace*{\algorithmicindent} \textbf{Input} Task description $T$, simulation API $e$

\begin{algorithmic}
%
\State \textbf{Step 1:} Inference
\State Success condition $sc \gets $ LLM$_1(T,e)$
\State Fitness functions $\{f_i\}_{i\in \mathcal{I}} \gets $ LLM$_2(T,e)$ (sample w/o rep.)
\State Timestamps $\{h_j\}_{j\in \mathcal{J}} \gets $ LLM$_3(T,e)$ (sample w/o rep.)
\State Measured quantities $\{g_k\}_{k\in \mathcal{K}} \gets $ LLM$_4(T,e)$ (sample w/o rep.)
\State BD mesh $\phi \gets \{g_k\circ h_j\}_{k\in \mathcal{K}, j\in \mathcal{J}}$ 
\State \textbf{Step 2:} Fitness exploration
\For{All fitnesses $\{f_i\}_{i\in\mathcal{I}}$}
\State Archives $S_{i} \gets $ Multi$\_$BD$\_$MES$_{b_1}(f_i,\phi)$
\EndFor
\State \textbf{Step 3:} Fitness selection
\State $\{s_i\}_{i\in \mathcal{I}} \gets \{$ size$(S_{i})\}_{i\in \mathcal{I}}$
\State $i_{max} \gets \underset{i}{\text{argmax}}\{s_i\}_{i\in \mathcal{I}}$
\State \textbf{Step 4:} Archive generation
\State Archive $S \gets $ Multi$\_$BD$\_$MES$_{b_2}(f_{i_{max}},\phi)$
\State \textbf{Return} S
\end{algorithmic}
\end{algorithm}

\section{Experiments}\label{sec:exp}
In this section, we detail the implementation of Algorithm \ref{alg:algo} used in our experiments and describe the experimental setting used to validate our method.

\subsection{Implementation: Candidate solution template}
In this paper, we choose the candidate solution template as a 6-dof position (the robot's initial position and orientation) and a 3-dof motion vector. After being initialized at its starting pose, the robot's gripper closes with the robot remaining immobile. Then, the robot carries out a linear motion in cartesian space following the motion vector. During this motion, the gripper keeps a constant orientation. The mutation operator is gaussian (with std $0.15$ for the initial xyz, $0.3$ for the initial rpy).\\
The initial 3D pose is contained in a cube with $20$ cm sides, centered around the object. We select this area such that it is small enough to ensure that the robot is likely to interact with the object, but large enough to avoid sampling many positions where the robot interpenetrates with the object. The initial gripper rotation $R$ is sampled from the half-sphere where the gripper points 'down', that is, $R \in SO(3): (Ru).u \leq 0$ with $u := (0,0,1)$ (in genesis, the gripper points upwards at its default orientation, hence the sign of the inequality). Finally, the motion vector has an amplitude between $5$ and $20$ cm, and has a positive vertical component (that is, the robot cannot move down). We choose these parameters to avoid motions that would be physically unrealistic or overly likely to produce unwanted collisions.

\subsection{Implementation: perception API}
The following information is available to the LLM: (i) the object base's position and orientation (we choose Euler angles as they are easier to interpret for a non-expert system such as an LLM); (ii) the contacts between the table and the robot (resp. object), (iii) the contacts between the robot's gripper and each link of the object (or object base for non-articulated objects) (iv) the robot gripper's position and orientation and (v) for articulated objects, the position (angle or prismatic displacement) of each joint.

\subsection{Environment and tasks}
\begin{figure}
    \centering
    \includegraphics[width=0.4\linewidth]{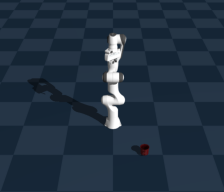} \includegraphics[width=0.4\linewidth]{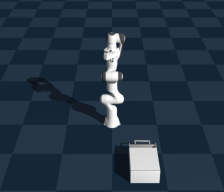}\vspace{0.1cm}\\
    \includegraphics[width=0.4\linewidth]{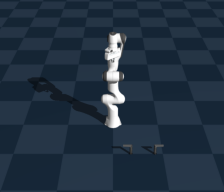} \includegraphics[width=0.4\linewidth]{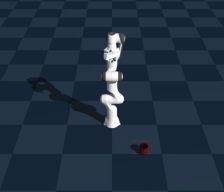}
    \caption{Example environment for each task in the genesis simulator. In reading order: grasping task, drawer task, faucet task, push task (table dimensions materialized by bright lines)}
    \label{fig:tasks}
\end{figure}

We validate our algorithm on $4$ simulated manipulation tasks. The archives obtained with Algorithm \ref{alg:algo} and each baseline are evaluated with an ground-truth success condition (unseen at inference time), and duplicate individuals (if any) are removed. The tasks, for which example environments are illustrated in Figure \ref{fig:tasks}; are the following:
\subsubsection{Grasp the mug}
\noindent \textbf{Expert success condition:} The object must touch the gripper and not the table; \textbf{Ground-truth BD:} Gripper xyz position at first contact with the object.
\subsubsection{Turn on the hot water}
\noindent \textbf{Expert success condition:} The hot water tap must be rotated at least $0.314$ radians; \textbf{Ground-truth BD:} Hot water tap opening at the end of the episode.
\subsubsection{Open the drawer at least 10cm}
\noindent \textbf{Expert success condition:} The prismatic joint between the drawer and the body must be extended at least 10cm past closed position; \textbf{Ground-truth BD:} Initial gripper xyz position
\subsubsection{Slide the mug on the table at least 5cm}
\noindent \textbf{Expert success condition:} The mug must be on the table, at least 5cm away from its starting position. \textbf{Ground-truth BD:} Object xy position at the end of the episode.

\subsection{Baselines}
We compare our method to the following baselines, all compute-matched (same population size, evaluation count):\\
\noindent \textbf{Naive inference:} Standard MES algorithm with success condition, fitness and BD inferred using an LLM (without our exploration procedure).\\
\noindent \textbf{Hand-crafted MES:} Standard MES algorithm with expert-written success condition, fitness and BD. The BD we use here corresponds to the ground-truth BD described above, and is typically a \textit{describing} BD \cite{QDG}, measuring diversity in task approach.\\
\noindent \textbf{Generic genetic algorithm (GA):} A standard genetic algorithm. We infer the fitness function with an LLM.\\
Since using multiple BD signals for QD-based manipulation learning is, to the best of our knowledge, yet unexplored, we do not evaluate a multi-BD hand-crafted baseline, as we lack best practices to design such an algorithm.

\subsection{Metrics}
\noindent \textbf{Success archive size:} We measure the number of successful individuals (measured with an expert-written success condition) output by each run.\\
\noindent \textbf{Ground-truth BD coverage:} 
For each task, we project the output archive from each baseline into the ground-truth (GT) BD space used for the hand-crafted baseline. This GT BD space is hand-crafted to capture meaningful diversity in task approaches, providing a measure of actual solution diversity.\\

\section{Results and discussion}\label{sec:results}
\begin{figure}
    \centering
    \includegraphics[width=0.35\linewidth]{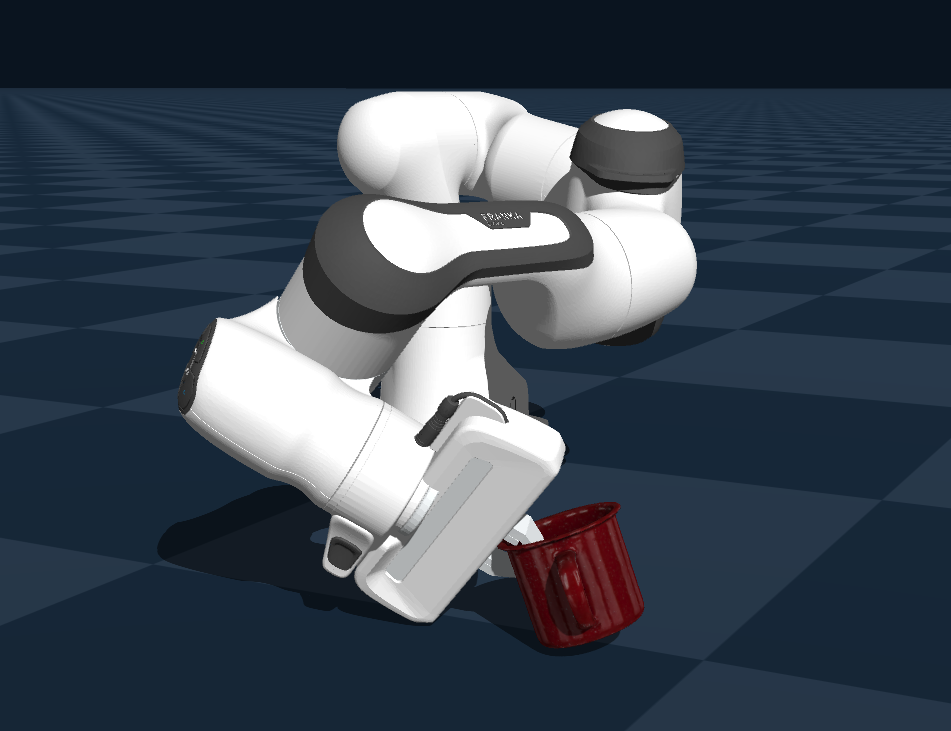} \includegraphics[width=0.35\linewidth]{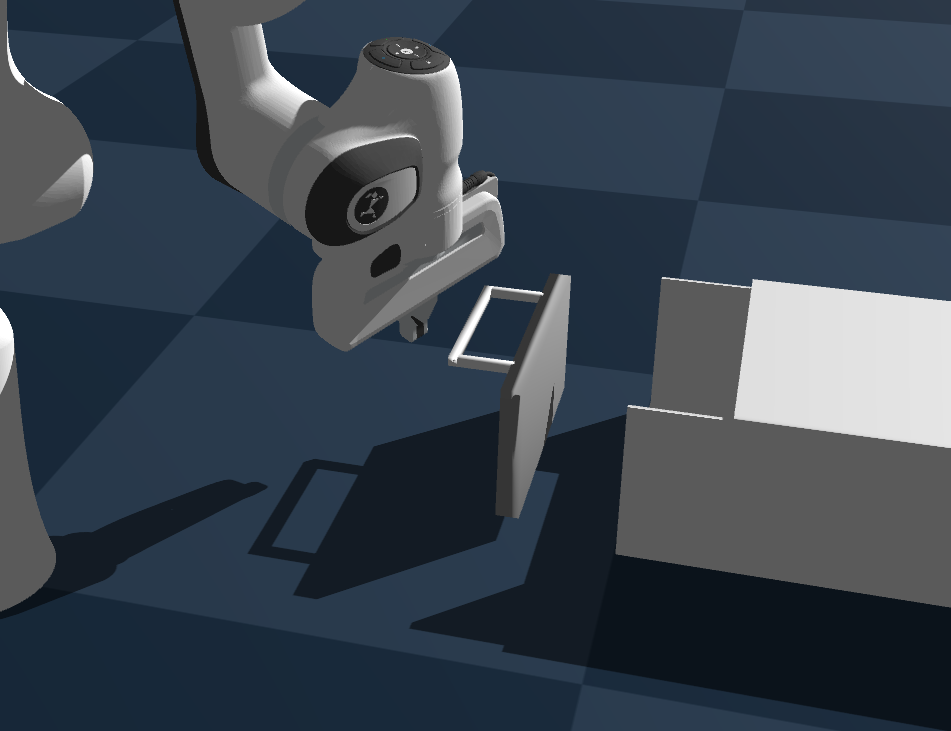}\vspace{0.1cm}\\
    \includegraphics[width=0.35\linewidth]{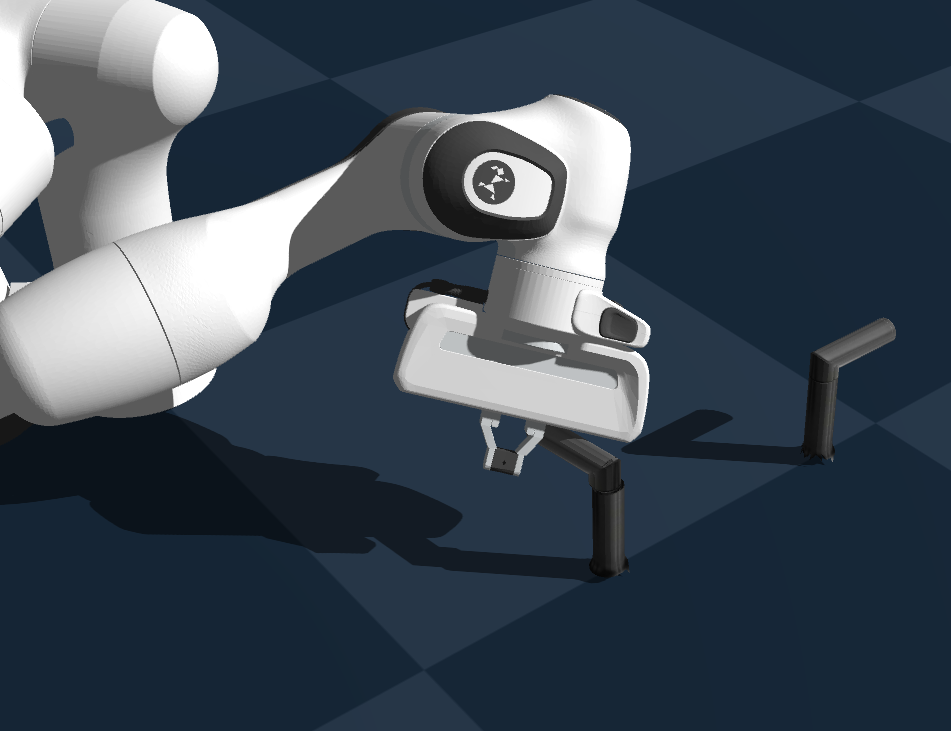} \includegraphics[width=0.35\linewidth]{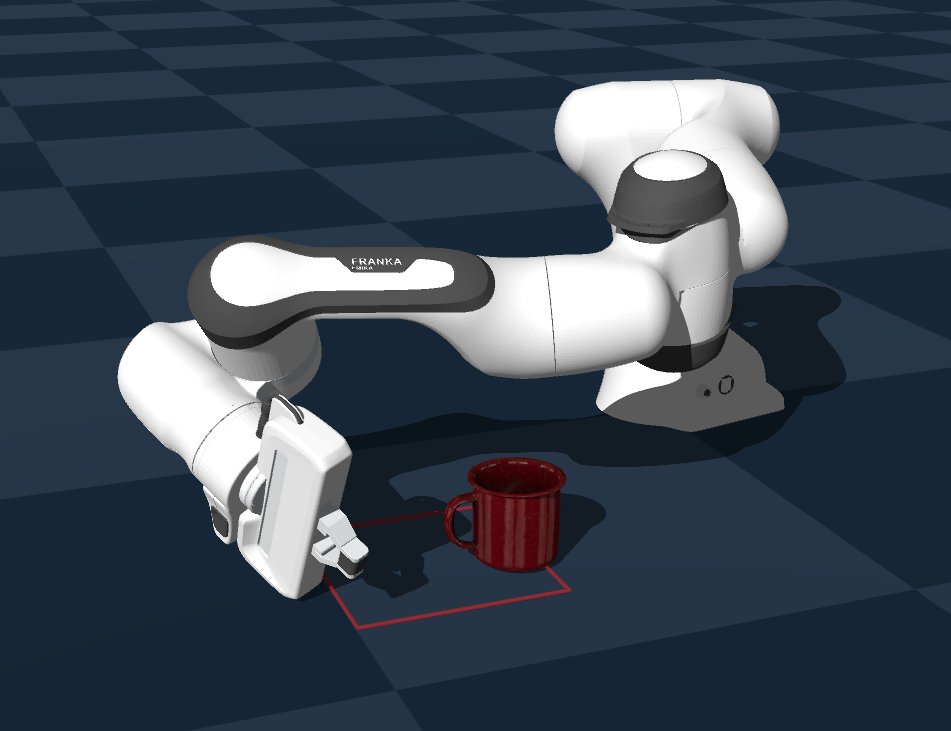}
    \caption{Example rollout for each task.}
    \label{fig:examples}
\end{figure}

For each inference step, we use GPT-4o with a temperature of $0$. The simulations are run using genesis \cite{genesis}, where GPU parallelization allows us to evaluate individuals rapidly. For each method, we attempt each task $5$ times. The metrics are given in $mean\pm std$ format in Tables \ref{tab:grasp_results} to \ref{tab:push_results}. Example rollouts of each task are given in Figure \ref{fig:examples}. On average, the \textbf{Naive inference} baseline finds significantly less solutions than our method, due to the inferred fitness and BD components being less adapted to the task. Interestingly, the \textbf{hand-crafted} baseline with GT fitness and BD also performs worse than our algorithm. We generally used \textit{describing} BDs \cite{qd1} as the ground-truth BDs. These are BDs that measure how functionally different from each other approaches are, and they tend to be quite sparse. In turn, this leads to small archives which struggle to generate successes. Inferred BDs are typically \textit{driving} BDs, meaning they are denser and can lead to many successful individuals early-on. With longer experiments (in this paper, we use $50$ generations per experiment), we expect that the hand-crafted baseline would eventually outperform the inferred baseline in GT BD coverage. Interestingly, maintaining several streams of diversity through a multi-BD approach allows us to get the best of both worlds, as grids associated with driving BDs rapidly fill up, while diversity in the GT BD dimension is still maintained. Further, while previous works \cite{qd2} rely on strong priors to overcome the sparsity of describing BDs, these are not needed with the multi-BD approach, providing encouraging insights into the openness of the method. While they are out of the scope of this paper, adaptations to the QD algorithm, such as injected random solutions in the archives, can be envisioned to improve the method's consistency, reducing its currently high variance across trials.\\
Note that the genetic algorithm baseline performs very poorly across all tasks, except the easier faucet task. This is unsurprising - evolutionary algorithms without diversity metrics perform poorly in sparse interaction settings - and shows the advantage of tackling the harder QD problem for robot manipulation. An example of the inferred BD mesh for the grasping task is given in Appendix \ref{sec:ap2}. For other examples, we refer to our Github.

\begin{table}[]
    \centering
    \begin{tabular}{|l|c|c|}
    \hline
     & Success archive size & GT BD size \\ \hline
     Hand-crafted & $5.8\pm 2.9$  & $5.8\pm 2.9$ \\
     Ours & $ 403.3 \pm 218.0 $  &  $ 375.5 \pm 180.5 $\\
     Direct inference  &$ 144.7\pm 15.9$  &$ 140.5\pm 12.3$ \\
     GA & $0\pm 0$ & $0\pm 0$ \\
    \hline
    \end{tabular}
    \caption{\textbf{Grasping task:} On this hard exploration problem, our approach finds more than twice the amount of grasps found through naive inference, and strongly outperforms hand-crafted MES.}
    \label{tab:grasp_results}
\end{table}

\begin{table}[]
    \centering
    \begin{tabular}{|l|c|c|}
    \hline
     & Success archive size & GT BD size \\ \hline
     Hand-crafted & $113.2 \pm 15.0$  & $113.2 \pm 15.0$ \\
     Ours & $ 2720.4 \pm 1245.6 $ & $334.4\pm 154.1$ \\
     Direct inference  & $110.0\pm8.64$ & $97.0\pm 13.6$ \\
     GA & $0.6 \pm 0.5$ & $0.6 \pm 0.5$ \\
    \hline
    \end{tabular}
    \caption{\textbf{Drawer task:} As above, our method outperforms baselines in ground-truth BD exploration and overall archive size.}
    \label{tab:drawer_results}
\end{table}

\begin{table}[]
    \centering
    \begin{tabular}{|l|c|c|}
    \hline
     & Success archive size & GT BD size \\ \hline
     Hand-crafted & $ 85.0\pm 0$ & $ 85.0\pm 0$\\
     Ours & $ 13492.2\pm 1042.5$ & $85.0\pm 0$ \\
     Direct inference  & $ 7932.0\pm 3780.1$ & $ 85.0\pm 0$\\
     GA & $525.6 \pm 19.4$ & $84.6\pm0.5$. \\
    \hline
    \end{tabular}
    \caption{\textbf{Faucet task:} On this easy exploration problem, all methods fill the GT BD space. However, our method finds significantly more trajectories.}
    \label{tab:faucet_results}
\end{table}

\begin{table}[]
    \centering
    \begin{tabular}{|l|c|c|}
    \hline
     & Success archive size & GT BD size \\ \hline
     Hand-crafted & $ 109.8\pm 1.6$  & $109.8\pm 1.6$ \\
     Ours & $85518.0 \pm 891.9$ & $ 140.6\pm5.7$ \\
     Direct inference  & $4028.2 \pm 2052.1 $ & $137.6\pm4.1$ \\
     GA & $455.2\pm14.5$& $8.6\pm1.9$ \\
    \hline
    \end{tabular}
    \caption{\textbf{Push task:} Our method again outperforms baselines. While the GT BD space is easily explored, our method again finds significantly more solutions. While the stock genetic algorithms finds many individuals, its coverage of the GT BD space is extremely poor.}
    \label{tab:push_results}
\end{table}

\section{Concluding remarks and future work}
In this work, we propose a technique to explore the BD and fitness spaces in QD algorithms for robotic manipulation using LLMs, autonomously solving tasks while only needing a free-form instruction describing the task. Then, we show how the BD mesh obtained from this exploration technique can be leveraged by a multi-BD variant of the MES algorithm. Through numerical experiments on a set of $4$ common manipulation tasks, we show how our method effectively generates diverse motion primitive archives and outperforms baselines, while requiring significantly less effort to parametrize than standard QD algorithms. Specifically, our technique generates large datasets in hard exploration problems by leveraging heterogeneous diversity metrics to rapidly generate individuals. In easy exploration problems, these multiple diversity streams also lead to large archives of motion primitives being found.\\
We envision two main directions for future work. The first is the integration of curriculum-based aspects to the learning pipeline. Complex motions involve prohibitively high-dimensional search spaces and complex evaluation procedures, and we foresee that combining simpler skills is the best way of tractably learning these complex motions. We envision QD algorithms being well-suited for such approaches, with the diverse trajectories facilitating skill chaining. Finally, automatically generating candidate solution templates tailored to each task (by adapting the number of waypoints, the interpolation procedure between each one of them and/or the gripper closing/opening synergies) and introducing priors (e.g. on the gripper's initial position and/or direction of motion) is a crucial next step to increase the diversity of approachable tasks, which will allow us to apply our method on more diverse manipulation and/or locomotion tasks. Tailoring the motion template, along with adapting domain randomization techniques to our setting, will also strongly improve the sim2real performance of our method, and we leave a thorough evaluation of the transferability of language-driven QD to future work.

\section*{Acknowledgements}
This work was supported by the EU Horizon Europe Framework Programme, through the PILLAR and euROBIN projects (grant agreements 101070381 and 101070596) and by PostGenAI@Paris ANR-23-IACL-0007 (France 2030).

\bibliographystyle{IEEEtran}
\bibliography{sample}

@inproceedings{cap,
  title={Code as policies: Language model programs for embodied control},
  author={Liang, Jacky and Huang, Wenlong and Xia, Fei and Xu, Peng and Hausman, Karol and Ichter, Brian and Florence, Pete and Zeng, Andy},
  booktitle={2023 IEEE International Conference on Robotics and Automation (ICRA)},
  pages={9493--9500},
  year={2023},
  organization={IEEE}
}

@article{llmpp,
  title={Llm+ p: Empowering large language models with optimal planning proficiency},
  author={Liu, Bo and Jiang, Yuqian and Zhang, Xiaohan and Liu, Qiang and Zhang, Shiqi and Biswas, Joydeep and Stone, Peter},
  journal={arXiv preprint arXiv:2304.11477},
  year={2023}
}

@article{QDG,
  title={Quality diversity under sparse reward and sparse interaction: Application to grasping in robotics},
  author={Huber, Johann and H{\'e}l{\'e}non, Fran{\c{c}}ois and Coninx, Miranda and Amar, F Ben and Doncieux, St{\'e}phane},
  journal={arXiv preprint arXiv:2308.05483},
  year={2023}
}

@article{eureka,
  title   = {Eureka: Human-Level Reward Design via Coding Large Language Models},
  author  = {Yecheng Jason Ma and William Liang and Guanzhi Wang and De-An Huang and Osbert Bastani and Dinesh Jayaraman and Yuke Zhu and Linxi Fan and Anima Anandkumar},
  year    = {2023},
  journal = {arXiv preprint arXiv: Arxiv-2310.12931}
}

@misc{VQE,
      title={Vector Quantized-Elites: Unsupervised and Problem-Agnostic Quality-Diversity Optimization}, 
      author={Constantinos Tsakonas and Konstantinos Chatzilygeroudis},
      year={2025},
      eprint={2504.08057},
      archivePrefix={arXiv},
      primaryClass={cs.NE},
}

@article{affordance,
  title={Task-Aware Robotic Grasping by evaluating Quality Diversity Solutions through Foundation Models},
  author={Appius, Aurel X and Garrabe, Emiland and Helenon, Francois and Khoramshahi, Mahdi and Chetouani, Mohamed and Doncieux, Stephane},
  journal={arXiv preprint arXiv:2411.14917},
  year={2024}
}

@article{cully2015robots,
  title={Robots that can adapt like animals},
  author={Cully, Antoine and Clune, Jeff and Tarapore, Danesh and Mouret, Jean-Baptiste},
  journal={Nature},
  volume={521},
  number={7553},
  pages={503--507},
  year={2015},
  publisher={Nature Publishing Group UK London}
}

@inproceedings{policy_me,
  title={Policy gradient assisted map-elites},
  author={Nilsson, Olle and Cully, Antoine},
  booktitle={Proceedings of the Genetic and Evolutionary Computation Conference},
  pages={866--875},
  year={2021}
}

@inproceedings{policy_sigaud,
  title={Diversity policy gradient for sample efficient quality-diversity optimization},
  author={Pierrot, Thomas and Mac{\'e}, Valentin and Chalumeau, Felix and Flajolet, Arthur and Cideron, Geoffrey and Beguir, Karim and Cully, Antoine and Sigaud, Olivier and Perrin-Gilbert, Nicolas},
  booktitle={Proceedings of the Genetic and Evolutionary Computation Conference},
  pages={1075--1083},
  year={2022}
}

@inproceedings{distill1,
  title={Map-elites with descriptor-conditioned gradients and archive distillation into a single policy},
  author={Faldor, Maxence and Chalumeau, F{\'e}lix and Flageat, Manon and Cully, Antoine},
  booktitle={Proceedings of the Genetic and Evolutionary Computation Conference},
  pages={138--146},
  year={2023}
}

@article{distill2,
  title={Generating behaviorally diverse policies with latent diffusion models},
  author={Hegde, Shashank and Batra, Sumeet and Zentner, KR and Sukhatme, Gaurav},
  journal={Advances in Neural Information Processing Systems},
  volume={36},
  pages={7541--7554},
  year={2023}
}

@misc{qd1,
      title={Quality Diversity under Sparse Reward and Sparse Interaction: Application to Grasping in Robotics}, 
      author={J. Huber and F. Hélénon and M. Coninx and F. Ben Amar and S. Doncieux},
      year={2023},
      eprint={2308.05483},
      archivePrefix={arXiv},
      primaryClass={cs.RO},
}

@misc{qd2,
      title={Speeding up 6-DoF Grasp Sampling with Quality-Diversity}, 
      author={Johann Huber and François Hélénon and Mathilde Kappel and Elie Chelly and Mahdi Khoramshahi and Faïz Ben Amar and Stéphane Doncieux},
      year={2024},
      eprint={2403.06173},
      archivePrefix={arXiv},
      primaryClass={cs.RO},
}

@misc{qd3,
      title={QDGset: A Large Scale Grasping Dataset Generated with Quality-Diversity}, 
      author={Johann Huber and François Hélénon and Mathilde Kappel and Ignacio de Loyola Páez-Ubieta and Santiago T. Puente and Pablo Gil and Faïz Ben Amar and Stéphane Doncieux},
      year={2024},
      eprint={2410.02319},
      archivePrefix={arXiv},
      primaryClass={cs.RO},
}

@article{l2r,
  title={Language to rewards for robotic skill synthesis},
  author={Yu, Wenhao and Gileadi, Nimrod and Fu, Chuyuan and Kirmani, Sean and Lee, Kuang-Huei and Arenas, Montse Gonzalez and Chiang, Hao-Tien Lewis and Erez, Tom and Hasenclever, Leonard and Humplik, Jan and others},
  journal={arXiv preprint arXiv:2306.08647},
  year={2023}
}

@article{kwon2023reward,
  title={Reward design with language models},
  author={Kwon, Minae and Xie, Sang Michael and Bullard, Kalesha and Sadigh, Dorsa},
  journal={arXiv preprint arXiv:2303.00001},
  year={2023}
}

@article{NS,
  title={Abandoning objectives: Evolution through the search for novelty alone},
  author={Lehman, Joel and Stanley, Kenneth O},
  journal={Evolutionary computation},
  volume={19},
  number={2},
  pages={189--223},
  year={2011},
  publisher={MIT Press}
}

@misc{card,
      title={A Large Language Model-Driven Reward Design Framework via Dynamic Feedback for Reinforcement Learning}, 
      author={Shengjie Sun and Runze Liu and Jiafei Lyu and Jing-Wen Yang and Liangpeng Zhang and Xiu Li},
      year={2024},
      eprint={2410.14660},
      archivePrefix={arXiv},
      primaryClass={cs.LG}, 
}

@misc{zeng2024learningrewardrobotskills,
      title={Learning Reward for Robot Skills Using Large Language Models via Self-Alignment}, 
      author={Yuwei Zeng and Yao Mu and Lin Shao},
      year={2024},
      eprint={2405.07162},
      archivePrefix={arXiv},
      primaryClass={cs.RO},
}

@article{targ,
  title={{TARG}: Tree of Action-reward Generation With Large Language Model for Cabinet Opening Using Manipulator},
  author={Park, Sung-Gil and Kim, Han-Byeol and Lee, Yong-Jun and Ahn, Woo-Jin and Lim, Myo Taeg},
  journal={International Journal of Control, Automation and Systems},
  volume={23},
  number={2},
  pages={449--458},
  year={2025},
  publisher={Springer}
}

@misc{supddown,
      title={Scaling Up and Distilling Down: Language-Guided Robot Skill Acquisition}, 
      author={Huy Ha and Pete Florence and Shuran Song},
      year={2023},
      eprint={2307.14535},
      archivePrefix={arXiv},
      primaryClass={cs.RO}, 
}

@misc{zeroshot,
      title={Zero-Shot Robotic Manipulation with Pretrained Image-Editing Diffusion Models}, 
      author={Kevin Black and Mitsuhiko Nakamoto and Pranav Atreya and Homer Walke and Chelsea Finn and Aviral Kumar and Sergey Levine},
      year={2023},
      eprint={2310.10639},
      archivePrefix={arXiv},
      primaryClass={cs.RO},
}

@INPROCEEDINGS{cotimg,
  author={Ni, Fei and Hao, Jianye and Wu, Shiguang and Kou, Longxin and Liu, Jiashun and Zheng, Yan and Wang, Bin and Zhuang, Yuzheng},
  booktitle={2024 IEEE/CVF Conference on Computer Vision and Pattern Recognition (CVPR)}, 
  title={Generate Subgoal Images Before Act: Unlocking the Chain-of-Thought Reasoning in Diffusion Model for Robot Manipulation with Multimodal Prompts}, 
  year={2024},
  volume={},
  number={},
  pages={13991-14000},
  doi={10.1109/CVPR52733.2024.01327}}

@misc{surfer,
      title={Surfer: Progressive Reasoning with World Models for Robotic Manipulation}, 
      author={Pengzhen Ren and Kaidong Zhang and Hetao Zheng and Zixuan Li and Yuhang Wen and Fengda Zhu and Mas Ma and Xiaodan Liang},
      year={2024},
      eprint={2306.11335},
      archivePrefix={arXiv},
      primaryClass={cs.RO},
}

@misc{genai,
      title={Generative Artificial Intelligence in Robotic Manipulation: A Survey}, 
      author={Kun Zhang and Peng Yun and Jun Cen and Junhao Cai and Didi Zhu and Hangjie Yuan and Chao Zhao and Tao Feng and Michael Yu Wang and Qifeng Chen and Jia Pan and Wei Zhang and Bo Yang and Hua Chen},
      year={2025},
      eprint={2503.03464},
      archivePrefix={arXiv},
      primaryClass={cs.RO},
}

@misc{genesis,
  author = {{Genesis} Authors},
  title = {Genesis: A Generative and Universal Physics Engine for Robotics and Beyond},
  month = {December},
  year = {2024},
  url = {https://github.com/Genesis-Embodied-AI/Genesis}
}

@misc{groot,
      title={GR00T N1: An Open Foundation Model for Generalist Humanoid Robots}, 
      author={{NVIDIA Gr00t team}},
      year={2025},
      eprint={2503.14734},
      archivePrefix={arXiv},
      primaryClass={cs.RO},
}

@misc{anytask,
      title={AnyTask: an Automated Task and Data Generation Framework for Advancing Sim-to-Real Policy Learning}, 
      author={Ran Gong and Xiaohan Zhang and Jinghuan Shang and Maria Vittoria Minniti and Jigarkumar Patel and Valerio Pepe and Riedana Yan and Ahmet Gundogdu and Ivan Kapelyukh and Ali Abbas and Xiaoqiang Yan and Harsh Patel and Laura Herlant and Karl Schmeckpeper},
      year={2026},
      eprint={2512.17853},
      archivePrefix={arXiv},
      primaryClass={cs.RO}, 
}

@article{ME,
  title={Illuminating search spaces by mapping elites},
  author={Mouret, Jean-Baptiste and Clune, Jeff},
  journal={arXiv preprint arXiv:1504.04909},
  year={2015}
}

@article{ME_nature,
  title={Robots that can adapt like animals},
  author={Cully, Antoine and Clune, Jeff and Tarapore, Danesh and Mouret, Jean-Baptiste},
  journal={Nature},
  volume={521},
  number={7553},
  pages={503--507},
  year={2015},
  publisher={Nature Publishing Group UK London}
}

@inproceedings{morel2022automatic,
  title={Automatic acquisition of a repertoire of diverse grasping trajectories through behavior shaping and novelty search},
  author={Morel, Aur{\'e}lien and Kunimoto, Yakumo and Coninx, Alex and Doncieux, St{\'e}phane},
  booktitle={2022 International Conference on Robotics and Automation (ICRA)},
  pages={755--761},
  year={2022},
  organization={IEEE}
}

@article{dexg,
  title={DexEvolve: Evolutionary Optimization for Robust and Diverse Dexterous Grasp Synthesis},
  author={Zurbr{\"u}gg, Ren{\'e} and Cramariuc, Andrei and Hutter, Marco},
  journal={arXiv preprint arXiv:2602.15201},
  year={2026}
}

@article{openvla,
  title={Openvla: An open-source vision-language-action model},
  author={Kim, Moo Jin and Pertsch, Karl and Karamcheti, Siddharth and Xiao, Ted and Balakrishna, Ashwin and Nair, Suraj and Rafailov, Rafael and Foster, Ethan and Lam, Grace and Sanketi, Pannag and others},
  journal={arXiv preprint arXiv:2406.09246},
  year={2024}
}

@article{smolvla,
  title={Smolvla: A vision-language-action model for affordable and efficient robotics},
  author={Shukor, Mustafa and Aubakirova, Dana and Capuano, Francesco and Kooijmans, Pepijn and Palma, Steven and Zouitine, Adil and Aractingi, Michel and Pascal, Caroline and Russi, Martino and Marafioti, Andres and others},
  journal={arXiv preprint arXiv:2506.01844},
  year={2025}
}

\begin{appendices}

\section{Prompts}\label{sec:ap1}
We report both BD sampling prompts (recall we separately sample for $h$ and $g$) for our method. The fitness prompt, omitted due to space constraints, is similar.\\

\textbf{BD inference first prompt:} (sampling for $h$)
\begin{spverbatim}
You are in charge of a robot skill acquisition pipeline. The goal is to learn simple manipulation skills using a quality-diversity approach. The robot arm ends in a parallel gripper, and the goal is to learn trajectories in joint space. The gripper is approximately 15cm long from the tip of the fingers to its center of mass.

The learning pipeline will be based on the genesis simulator. The scene will contain the robot and relevant objects and/or furniture for the task.

In quality-diversity, a fitness component is necessary to evaluate candidate solutions, and an additional quantity, called the behavior descriptor, is used to ensure solutions are both high-performing and diverse. Ideally, optimizing behavioral coverage should lead to an archive of diverse, high-performing solutions.

Here is the task: "TASK"

You will be given access to functions measuring select physical quantities from the simulator. Each candidate solution will be rendered in simulation, and a relevant information will be stored in a list called infos. Each element of the list infos[k] is a dictionary with the quantities of interest. Here are the available elements, with a high-level description: 

{PERCEPTION API}
    
The goal is to design a behavior descriptor function. Candidate solutions will be considered diverse if their behavior descriptors are different from each other. This diversity should be used to promote different approaches and encourage exploration in behavioral space. However, the descriptor doesn't need to measure a candidate solution's performance, as this will be done later.

First, propose a set of 4 relevant timestamps within the simulation. Such timestamps can correspond to fixed moments in the simulation, for example the beginning or the end or interesting events such as the first contact between the gripper and the object/a part of the object. These timestamps will be applied to compute the BD.

Then, complete the following python functions that extract the information at the chosen timestamp out of the entire list of infos. These will be used to compute the BD later. If the desired timestamp is not contained in the information list, the functions should return None.

The code to fill out is:
{Function templates}

\end{spverbatim}

\textbf{BD inference second prompt:} (sampling for $g$)
\begin{spverbatim}
You are in charge of a robot skill acquisition pipeline. The goal is to learn simple manipulation skills using a quality-diversity approach. The robot arm ends in a parallel gripper, and the goal is to learn trajectories in joint space. The gripper is approximately 15cm long from the tip of the fingers to its center of mass.

The learning pipeline will be based on the genesis simulator. The scene will contain the robot arm and relevant objects and/or furniture for the task.

In quality-diversity, a fitness component is necessary to evaluate candidate solutions, and an additional quantity, called the behavior descriptor, is used to ensure solutions are both high-performing and diverse. Ideally, optimizing behavioral coverage should lead to an archive of diverse, high-performing solutions.

Here is the task: "TASK"

You will be given access to functions measuring select physical quantities from the simulator. Each candidate solution will be rendered in simulation, and a relevant information will be stored. The most interesting timestep within the simulation will be automatically selected, and you will have access to a dictionary with the quantities of interest at this timestep. Here are the available elements, with a high-level description: 

{PERCEPTION API}
    
The goal is to design a behavior descriptor function. Candidate solutions will be considered diverse if their behavior descriptors are different from each other. This diversity should be used to promote different approaches and encourage exploration in behavioral space. However, the descriptor doesn't need to measure a candidate solution's performance, as this will be done later.

Please make a list of 4 simple and interesting physical quantities, such as gripper or object position, to use for the BD. The quantities should, in general, be low-dimensional (from 2 to 5) tuples of real values. When using posiions or orientations, it is better to keep the 3D value, instead of keeping only some elements.

Then, please write the python functions extracting the desired quantities from the info dictionnary obtained above. The output of each python function should be a tuple of 1-dimensional values.

The code to fill out is:
{Function templates}
\end{spverbatim}

\section{Example inferred functions}\label{sec:ap2}
We give below an example of the Fitness and BD samples inferred for the "Grasp the mug" task.\\
\noindent \textbf{Fitness:} (averaged over simulation timesteps)
\begin{itemize}
    \item Object vertical position
    \item Negative (gripper-object) distance
    \item $1$ if the gripper is touching the object
\end{itemize}

\noindent \textbf{BD timesteps:} ($h$)
\begin{itemize}
    \item First simulation step
    \item First contact between the gripper and the mug
    \item First time step without mug-table contact
    \item Last time step
\end{itemize}

\noindent \textbf{BD information:} ($g$)
\begin{itemize}
    \item Gripper xyz position
    \item Gripper rpy orientation
    \item Mug xyz position
    \item $\{\mathds{1}_{\text{gripper-mug contact}},\mathds{1}_{\text{mug-floor contact}}\}$
\end{itemize}

\end{appendices}

\end{document}